\documentclass[conference]{IEEEtran}
\IEEEoverridecommandlockouts
\usepackage{cite}
\usepackage{amsmath,amssymb,amsfonts}
\usepackage{graphicx}
\usepackage{textcomp}
\usepackage{amsmath}
\usepackage{breqn}
\usepackage{xcolor}
\usepackage{multirow}
\def\BibTeX{{\rm B\kern-.05em{\sc i\kern-.025em b}\kern-.08em
    T\kern-.1667em\lower.7ex\hbox{E}\kern-.125emX}}

\usepackage{algorithm}
\usepackage{algcompatible} 

\algnewcommand\algorithmicreturn{\textbf{return}}
\algnewcommand\RETURN{\State \algorithmicreturn}%

\begin{document}

\raggedbottom

\title{Federated XGBoost on Sample-Wise Non-IID Data\\}

\author{
    \IEEEauthorblockN{Katelinh Jones}
    \IEEEauthorblockA{Pennsylvania State University \\
    State College, Pennsylvania \\
    kgj5045@psu.edu}
    \and
    \IEEEauthorblockN{Yuya Jeremy Ong}
    \IEEEauthorblockA{IBM Research - Almaden \\
    San Jose, California \\
    yuyajong@ibm.com}
    \and
    \IEEEauthorblockN{Yi Zhou}
    \IEEEauthorblockA{IBM Research - Almaden \\
    San Jose, California \\
    yi.zhou@ibm.com}
    \and
    \IEEEauthorblockN{Nathalie Baracaldo}
    \IEEEauthorblockA{IBM Research - Almaden \\
    San Jose, California \\
    baracald@us.ibm.com}
}

\maketitle

\begin{abstract}
Federated Learning (FL) is a paradigm for jointly training machine learning algorithms in a decentralized manner which allows for parties to communicate with an aggregator to create and train a model, without exposing the underlying raw data distribution of the local parties involved in the training process. Most research in FL has been focused on Neural Network-based approaches, however Tree-Based methods, such as XGBoost, have been underexplored in Federated Learning due to the challenges in overcoming the iterative and additive characteristics of the algorithm. Decision tree-based models, in particular XGBoost, can handle non-IID data, which is significant for algorithms used in Federated Learning frameworks since the underlying characteristics of the data are decentralized and have risks of being non-IID by nature. In this paper, we focus on investigating the effects of how Federated XGBoost is impacted by non-IID distributions by performing experiments on various sample size-based data skew scenarios and how these models perform under various non-IID scenarios. We conduct a set of extensive experiments across multiple different datasets and different data skew partitions. Our experimental results demonstrate that despite the various partition ratios, the performance of the models stayed consistent and performed close to or equally well against models that were trained in a centralized manner.
\end{abstract}


\section{Introduction}
The use of machine learning (ML) has amassed wide adoption across different domains and industries and has become a critical component of various processes such as finance, healthcare, and the sciences. However, such implementation of model training processes requires a significant amount of quality data sources, which are often fragmented across different tenants in a decentralized manner. Furthermore, many of the data available from such sources may include private or confidential data, such as Personal Identifiable Information (PII) and Personal Health Information (PHI), which often have strict data governance and privacy regulations around the use and portability of data sources. Examples of regulatory data policies include Health Insurance Portability and Accountability Act (HIPPA), the European General Data Protection Regulation (GDPR), and the California Consumer Privacy Act (CCPA). These regulatory practices have become critical requirements to ensure better stewardship of handling users' private data securely and responsibly.

To comply with these regulatory restrictions on the data, Federated Learning has emerged as a new paradigm for training machine learning models in a distributed manner while maintaining the privacy of the individual party's data \cite{kone2016}. Rather than training such machine learning models within a monolithic environment, training models in Federated Learning instead attempts to learn the parameters of the model using each party's underlying raw data distribution without directly sharing the data across the federation. Instead, the use of an \textit{aggregator} or a single machine responsible for fusing model parameters collected from a set of local \textit{parties} is used to train a single machine learning model based on the data from each of the local parties. Well-studied approaches for Federated Learning are focused primarily on the use of Neural Network-based approaches, such as Fed-Average \cite{mcmahan2016}. These methods primarily are based on the paradigm of each party training their models locally on their respective data, and accordingly sending their respective parameters to a centralized aggregator, where they perform some type of average fusion methods to aggregate the learned model results.

However, one particular type of models which have not been explored as much in the context of Federated Learning are Decision Tree-based models \cite{ong2022tree}, such as Random Forests \cite{truex2019hybrid} and Gradient Boosted Decision Trees \cite{ong2020adaptive}. Gradient boosted decision trees, e.g., XGBoost \cite{chen2015xgboost}, have recently been proposed as an ensemble tree-based model to improve the performance of CART decision trees. It utilizes a gradient boosting-based approach to optimize the tree split against a predefined loss function, e.g., mean-squared loss for regression problems and cross-entropy loss for classification problems, etc.

Within the context of Federated Learning, Gradient Boosted Decision Trees provide various key benefits compared to other model types \cite{ong2020adaptive}. First, these models handle missing data out of the box, which therefore does not require additional overhead for data preprocessing. Second, these models are based on decision-tree-based data structures which implicitly generate interpretable structures, making them inherently explainable out of the box. These models, therefore are highly desirable, especially in the context of Federated Learning, as they are often used within highly regulated environments where model governance and audits are often strictly required. Finally, Gradient Boosted Decision Tree models can handle non-IID, or \textit{independent and identically distributed}, data out of the box, making this a desirable property under scenarios where the data is distributed in a decentralized manner. In this paper, our focus is specifically on the third property of this algorithm where we subject the model to various non-IID scenarios and evaluate to see how well they perform under those conditions when trained in a Federated Learning setting.

In various works that have evaluated Federated Learning algorithms, it is often the case that these models are evaluated under scenarios where the data distributions of the local parties are assumed to be IID. However, it is often the case in practice that the data for each party are non-IID, as data distributions of each local party within the federation are stored in an environment where they are located in a disparate and fragmented manner. Such fragmentation in data sources can introduce potential risks for data distributions to differ completely due to factors such as geographical, cultural, and other organizational differences which can emerge from party to party. The result of such biases and imbalances emerging from the data distribution leads to various non-IID distributions such as sample/count-based, feature-based, and target ratio-based non-IID distributions. Therefore, it is paramount for Federated Learning algorithms to be able to handle data with non-IID properties to ensure that no such impact from bias and other influences on the data distribution degrades the performance of the model.

In this paper, we investigate the effects of training a Federated Learning-based Gradient Boosted Decision Tree algorithm on sample-based non-IID distribution. In particular, we focus on investigating the effects of how Federated Gradient Boosted Decision Tree-based models can handle various sample-wise non-IID data distributions. In particular, the objective of this study is to conduct an extensive set of experiments across various datasets, under different sample sizes, feature dimensions, and sample-wise ratios amongst other parties. Through these experiments, we uncover the strengths and limitations of our proposed Federated XGBoost algorithm to see how our method scales and performs under different sample-wise non-IID scenarios through a systematic set of evaluations. We demonstrate robust performance metrics across various learning tasks and show the efficacy of our proposed methods against various non-IID scenarios.

The rest of this paper is structured as follows. Section 2 introduces related work on Federated XGBoost and prior studies of the performance characteristics of Federated Learning algorithms under non-IID data scenarios. In Section 3, we present various preliminary concepts on Federated Learning, the fundamentals of XGBoost, and our proposed Federated XGBoost algorithm. In Section 4, we describe our experimental setup, and respectively in Section 5 we demonstrate our experimental results and analysis. Finally, Section 6 concludes our paper, as well as suggests directions for future research.

\section{Literature Review}
In this section, we first survey some of the proposed methods that have been published recently in the literature around tree-based methods for Federated Learning, in particular, Gradient Boosted Decision Trees. We briefly explore different dimensions of these tree-based approaches that have emerged in the literature, exploring aspects such as horizontal versus vertical Federated Learning and data privacy and security protection methodologies. We then also survey some previous works that have explored the training of Federated Learning algorithms over non-IID distributions and some of their early contributions to tree-based approaches.

\subsection{FL for Decision Tree-Based Models}
Originally, Federated Learning (FL) \cite{kone2016} emerged as a paradigm for collaboratively training a single global machine learning model without having to share or expose the local party's raw data distributions among the federation, thus providing a layer of security for each party that participates in the training process. Typically these algorithms are often implemented as linear-based models or Neural Network-based approaches. In recent years, the adoption of Decision Tree-based models for Federated Learning \cite{ong2022tree} has begun to emerge as a new type of model specifically for Federated Learning models, compared to traditional models such as linear models and neural network-based approaches. The majority of tree-based approaches proposed in Federated Learning are based on either a Random Forest (RF) approach, or a variant of the Gradient Boosted Decision Tree-based algorithm, such as XGBoost \cite{chen2015xgboost} or LightGBM \cite{ke2017lightgbm}.

\subsubsection{Horizontal vs Vertical FL}
One key dimension to consider in training Federated Learning algorithms is whether the data dimensions are shared feature-wise or sample-wise. In other words, we consider the distinction based on whether they can be used \textit{horizontally} or \textit{vertically}. Federated Learning algorithms can be categorized into two different types, \textit{horizontal} or \textit{vertical} Federated Learning, depending on what dimensions are common across the participants within the training of the model. In horizontal Federated Learning, parties share the same set of features, while in vertical Federated Learning, parties share the same set of data sample identifiers. Depending on how the data is structured across the party, the resulting communication topology of the federated learning system as well as the method and type of information exchanged can differ greatly.

The majority of approaches are based on horizontal data partitions. Relatively, very few methods such as SecureBoost \cite{cheng2019secureboost}, S-XGB and HESS-XGB \cite{Fang2020AHF}, and SecureGBM \cite{feng2019securegbm} are examples of Vertical Federated Learning architectures. In most vertical-based Federated Learning systems, feature alignment must be performed in some fashion before the model training process. However, given that the underlying data structure of the tree algorithm is based on finding the optimal feature to split on, the underlying data representation of the model is dependent on the availability of such feature existing in the local party's dataset.

\subsubsection{Data Privacy and Security Frameworks}
Applying federated learning under different scenarios may require substantially different data governance policies, which includes either \textit{Trusted Federations} or \textit{Protected Federations} \cite{ong2022tree}. 

Trusted Federations involve the use of dimensional reduction-based techniques which reduces the overall fidelity of the data to prevent other parties from having an actual view of the raw data - instead having some surrogate representation or approximation of the raw data. Examples of approaches that employ this method include: Li et. al \cite{li2020practical} which implements a Locality Sensitivity Hashing (LSH) method, Yang et. al \cite{yang2019tradeoff} proposed a clustering-based k-anonymity scheme, and Ong et. al \cite{ong2020adaptive} that utilizes a party-adaptive histogram approximation mechanism. This form of data obfuscation is slightly different from the additive form of statistical methods such as differential privacy where noise is added to the data, however, it follows a very similar principle for hiding the raw data distribution from potential adversaries.

On the other hand, Protected Federation policies involve the use of stronger security measures and defenses against potential data leaks and exposure of clients' data. These techniques prevent adversaries from obtaining direct unauthorized access to the raw data distribution of those participating in the joint training of a machine learning model or inference of private data. The two most common methods within the Protected Federation policies are \textit{statistical methods} and \textit{encryption-based methods}.

Statistical methods of privacy protection employ techniques such as k-anonymity \cite{sweeney2002k} and differential privacy (DP) \cite{dwork2006calibrating} which are two commonly employed techniques for tree-based Federated Learning methods. Examples of methods that employ these schemes include Yang et. al \cite{yang2019tradeoff}, Truex et. al \cite{truex2019hybrid}, Peltari \cite{pelttari2020federated}, Liu et. al \cite{liu2020federated}, and Tian et. al  \cite{tian2020federboost}. These methods define some statistical measures of privacy guarantees with an upper-bound error margin for some defined privacy budget. Although these methods do not require as much computational overhead compared to encryption-based methods, the major disadvantage of these schemes primarily degrades the overall accuracy performance of the machine learning task at hand due to additional additive errors introduced into the data. 

On the other hand, encryption-based approaches utilize encryption as a mechanism to perform data operations directly on encrypted data sources through methods such as Homomorphic Encryption (HE), Secret Sharing (SS), and Secure Multi-Party Computation (SMC).  Examples of encryption-based security protocols include: SecureBoost \cite{cheng2019secureboost}, FedXGB \cite{liu2019boosting}, and HESS-XGB \cite{Fang2020AHF} which employs homomorphic encryption. Alternatively, methods such as S-XGB \cite{Fang2020AHF} and PrivColl \cite{leung2020towards} use Secret Sharing. Here, the major trade-off to take into consideration is the additional computational and network communication overhead incurred during the process of encryption and decryption of data which increases the overall runtime necessary for training a model, in exchange for near lossless data being transmitted, which translates to better model performance.

In this paper, we focus on enterprise settings where collaborations do not require secure aggregation schemes. Our design does not utilize encryption-based methods, yielding significant training time advantages over some of the prior methods. Our proposed method, instead, utilizes trusted federation-based governance where we apply a histogram-based method to approximate the raw distribution of the data through a surrogate representation of the data. Therefore, parties do not need to reveal their exact local data distributions to others and can collaboratively grow the tree without exposing their raw data distribution amongst the federation.

\subsection{Non-IID Data in Federated Learning}
One of the largest challenges when training Federated Learning models involves utilizing data across heterogeneous sources, which often are non-IID. Many prior works have explored these areas of training Federated Learning algorithms under various non-IID scenarios. Zhao et. al \cite{zhao2018} explored the effects of non-IID data for Neural Network-based Federated Learning algorithms and demonstrated significant performance degradation of the model. Sattler et. al \cite{sattler2019robust} proposes a compression framework that allows for robustness against non-IID data when training models in Federated Learning. However, these approaches mainly evaluate Federated Learning models over Neural Network-based approaches. Fan et. al \cite{fan2020smart} on the other hand, propose a method for Gradient Boosted Decision Tree-based models for Federated Learning and evaluates the effects of skewed label data. However, their work does not perform a systematic evaluation of the effects of training a Federated Learning model over different types of sample-based non-IID data. In our work, we design very specific experiments that test the Federated XGBoost algorithm proposed by Ong et. al \cite{ong2020adaptive} under different non-IID scenarios to evaluate its performance over different types of data distribution skews in terms of sample size distributions. By studying how different allocation of sample sizes impacts the overall model performance, we can better evaluate the efficacy of Federated Gradient Boosted Decision Trees to see what the potential weakness or possible performance issues are that arise from these types of data distributions.
\section{Federated Gradient Boosting}
In this section, we introduce the underlying methodology for how to train a Federated Gradient Boosted Decision Tree algorithm. We first begin by discussing the necessary preliminaries including notations, background on Federated Learning, and Gradient Boosted Decision Trees. Then we introduce the method behind how Gradient Boosted Decision Trees are trained within a Federated Learning framework.

\subsection{Preliminaries}
\subsubsection{Basic Notations for FL Systems}

\begin{figure}[]
    \centering
    \includegraphics[width=0.98\columnwidth]{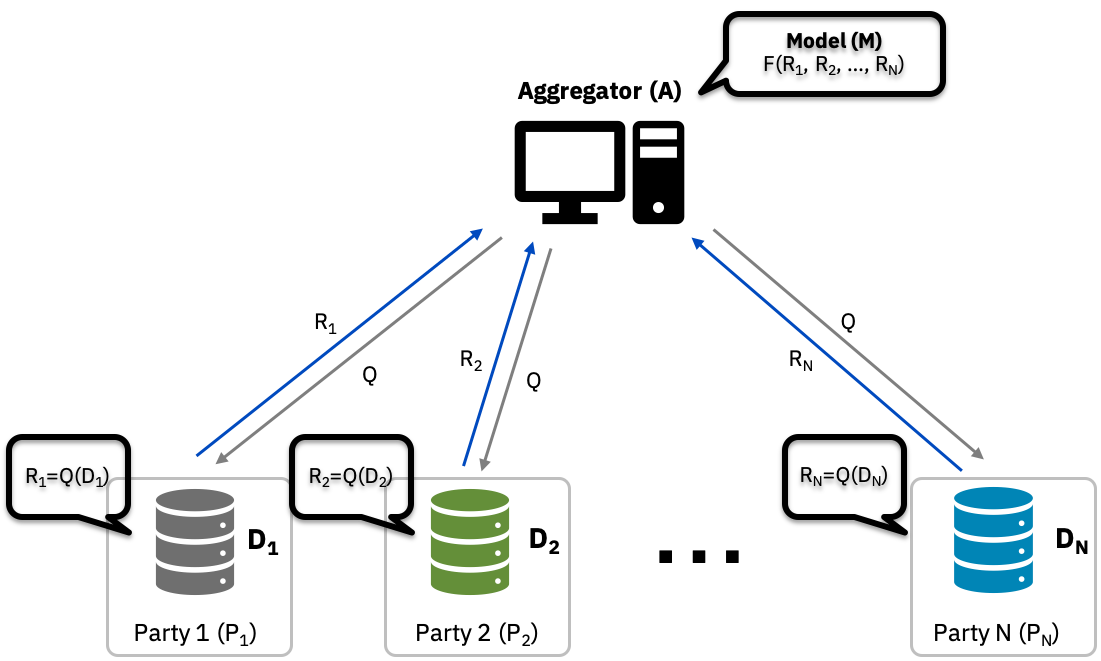}
    \caption{\footnotesize A general architecture of federated learning}
    \label{fig:fl-overview}
\end{figure}

We define a Federated Learning system based on a federation consisting of $n$ parties, $\mathcal{P} = \{P_1, P_2, ..., P_n\}$, with a disjoint dataset $\mathcal{D}_1, \mathcal{D}_2, ..., \mathcal{D}_n$ each sharing the same number of features (\textit{i.e. horizontal FL}), $m$, and an aggregator $\mathcal{A}$, which orchestrates the training procedure. Figure~\ref{fig:fl-overview}, demonstrates the high-level architecture of our Federated Learning setup, where for each training round, the aggregator issues a query $Q$ to the available parties in the system, and a party $\mathcal{P}_i$ replies to that query with $r_i$ based on a computed value from its local dataset $\mathcal{D}_i$ and replies to the aggregator. The aggregator then collects the responses from each party and fuses the responses to update the global ML model, defined by $\mathcal{M} = \mathcal{F}(r_1, r_2, ..., r_n)$ which resides on the aggregator side. This process gets performed for multiple iterations until some termination criteria, such as model convergence or a user-defined heuristic have been met.

\subsubsection{Gradient Boosted Decision Trees}
In this section, we now briefly introduce the \textbf{eXtreme Gradient Boosted (XGBoost)} algorithm, which is based on a highly optimized and efficient variant of the Gradient Boosted Decision Tree defined by Friedman et. al \cite{friedman2001greedy}. We refer the readers to Chen et. al's paper \cite{chen2015xgboost} for further details.

Given a dataset $\mathcal{D}$ with $n$ samples and $m$ features, $\mathcal{D}=\{(x_i, y_i)\}^{n}_{i=1}$, where $x_i \in \mathbb{R}^m$ and $y_i\in\mathbb{R}$, the predictions output from the XGBoost model, $\hat{y}_i$, is defined as a additive tree-based ensemble model, $\phi(x_i)$, comprising of $K$ additive functions, $f_k$, defined as:
$$\hat{y}_i = \phi(x_i) = \sum_{k=1}^{K} f_k(x_i), \; f_k \in \mathcal{F}$$

\noindent where $\mathcal{F} = \{f(x) = w_{q(x)}\}$ is a collection of Classification and Regression Trees (CART), such that the function $q(x)$ maps each input feature $x$ to one of $T$ leaves in the tree by a weight vector, $w \in \mathbb{R}^T$.

Given the defined model prediction above, the XGBoost algorithm minimizes the following regularized loss function:
$$\tilde{\mathcal{L}}=\sum_i l(y_i,\hat{y}_i) + \sum_k \Omega(f_k)$$

\noindent where $l(y_i,\hat{y}_i)$ is the loss function of the $i$th sample between the prediction $\hat{y}_i$ and the target value $y_i$, and $\Omega(f_k)=\gamma T + \frac{1}{2}\lambda\|w\|^2$ is the regularization component. This component discourages each $k^{th}$ tree, $f_k$, from over-fitting through hyperparameters  $\lambda$, the regularization parameter penalizing the weight vector $w$, and $\gamma$, a term penalizing the tree from growing additional leaves.

To approximate the loss function, a second-order Taylor expansion function is used, as defined by:
$$\mathcal{L}^{(t)} \simeq \sum_{i=1}^n [l(y_i,\hat{y_i}^{(t-1)})+g_if_t(x_i)+\frac{1}{2}h_if_t^2(x_i)] + \Omega(f_t)$$

\noindent As the tree is trained in a recursively additive manner, each iteration index of the training process is denoted as $t$, hence $\mathcal{L}^{(t)}$ denotes the $t^{th}$ loss of the training process. Here, we also define the gradient and the second order gradient, or the Hessian, respectively, as follows:
$$g_i = \partial_{\hat{y}_i^{(t-1)}}l(y_i, \hat{y}_i^{(t-1)})$$
$$h_i = \partial_{\hat{y}_i^{(t-1)}}^2 l(y_i, \hat{y}_i^{(t-1)})$$

Given the derived gradients and Hessians for a given $q(x)$, we can compute the optimal weights of leaf $j$ using:
$$w^*_j = -\frac{G_j}{H_j + \lambda}$$

\noindent where $G_j = \sum_{i \in I_j} g_i$ and $H_j = \sum_{i \in I_j} h_i$ are the total summation of the gradients and Hessians for each of the specific data sample indices, $I_j$, respectively. To efficiently compute the optimal weights $w^*_j$, we can greedily maximize the gain score to search for the best partition value for a leaf node at each iteration efficiently. This gain score is defined as follows:
$$Gain=\frac{1}{2}\left[\frac{G^2_L}{H_L + \lambda} + \frac{G^2_R}{H_R + \lambda} - \frac{(G_L+G_R)^2}{(H_L+H_R)+\lambda} \right] - \gamma$$

\noindent Here, $L$ and $R$ correspondingly consider the sum of the gradients and Hessians based on the specific index of the left and right children of the given leaf node, $I_L$, and $I_R$, respectively. \\

\subsection{Federated XGBoost}
\begin{algorithm}
\algdef{SE}[SUBALG]{INDENT}{ENDINDENT}{}{\algorithmicend\ }%
\algtext*{INDENT}
\algtext*{ENDINDENT}

\caption{Federated Gradient Boosted Decision Tree}
\label{alg:pax_algo}

\textbf{Input:} 
    $\mathcal{D}$, Input Dataset; $A$, Aggregator; 
    $P$, Participating Parties in FL Training; 
    $\delta$, Surrogate Representation Histogram Relative Error;
    $\epsilon$, Gradient Boosting Histogram Approximation Bin Size;
    $T$, Maximum Number of Training Rounds; 
    $l$, Model Loss Function \\
\textbf{Output:} $f^{(\mathcal{A})}$, Trained Global XGBoost Model

\begin{algorithmic}[1]
\STATE $\{(Y_1, n_1), ..., (Y_{\mathcal{P}}, n_{\mathcal{P}})\} \gets obtain\_lp\_target\_sum()$
\STATE $f^{(A)}_\emptyset \gets \dfrac{Y_1 + ... + Y_{\mathcal{P}}}{n_1 + ... + n_{\mathcal{P}}}$
\STATE

\FOR {$i=1, ..., |\mathcal{P}|$}
    \STATE $\tilde{\mathcal{D}}^{(p_i)}_X \gets construct\_histogram(\mathcal{D}^{(p_i)}_X, \delta)$
    \STATE $p_i$ Transmits $\tilde{\mathcal{D}}^{(p_i)}_X$ to $\mathcal{A}$
\ENDFOR
\STATE $\tilde{\mathcal{D}}^{\mathcal{A}}_X \gets merge\_histogram(\{ \tilde{\mathcal{D}}^{(1)}_X, ..., \tilde{\mathcal{D}}^{(P)}_X \})$
\STATE

\REPEAT
    \STATE ($G^{(\mathcal{A})}, H^{(\mathcal{A})}) \gets (\emptyset,  \emptyset$)
    \FOR {$i=1, ..., |\mathcal{P}|$}
        \STATE $\mathcal{A}$ Transmits $f^{(\mathcal{A})}_t$ to Party: $f^{(p_i)}_t \gets f^{(\mathcal{A})}_t$
        \STATE $p_i$ Generate Predictions: $\hat{y}_t^{(p_i)} = f_t^{(p_i)}(\tilde{\mathcal{D}}^{(p_i)}_X)$
        \STATE $p_i$ Computes $g^{(p_i)}$ and $h^{(p_i)}$
        \STATE $p_i$ Transmits $g^{(p_i)}$ and $h^{(p_i)}$ to $\mathcal{A}$
        \STATE $G^{(\mathcal{A})} \gets G^{(\mathcal{A})} \cup g^{(p_i)}$
        \STATE $H^{(\mathcal{A})} \gets H^{(\mathcal{A})} \cup h^{(p_i)}$
    \ENDFOR

    \STATE $G^{(\mathcal{A})}_m, H^{(\mathcal{A})}_m \gets merge\_hist(G^{(\mathcal{A})}, H^{(\mathcal{A})})$
    \STATE $f_t^{(\mathcal{A})} \gets grow\_tree(\tilde{\mathcal{D}}^{\mathcal{A}}_X, G^{(\mathcal{A})}_m, H^{(\mathcal{A})}_m, \epsilon)$
\UNTIL{$t \leq T$ or other termination criteria.}
\end{algorithmic}

\end{algorithm}

In this section, we introduce a slightly modified version of the XGBoost-based Federated Learning training method, also known as the \textit{Party-Adaptive XGBoost} (PAX), proposed by Ong et. al \cite{ong2020adaptive}. One of the key differences we introduce in this paper is around the methodology of selecting the bin sizes for the individual party's data, and how we simplify the implementation by utilizing a fixed bin size across all parties when generating the underlying histogram representation of the local party's data. As noted previously, one of the major challenges in training a gradient-boosted decision tree is to find the optimal feature and value to split using the computed gain score. Similar to the method of decomposing the routines of the algorithm as described in the previous section, the same principles of the aggregator querying each party for its distribution statistics, fusing those statistics, and finding the optimal partition based on those fused statistics holds for the Gradient Boosted Decision Tree. In this case, instead of simple data count statistics, we deal with histograms of data distributions instead.

Optimized GBDT methods such as XGBoost \cite{chen2015xgboost} and LightGBM \cite{ke2017lightgbm} utilize a \textit{quantile-based approximation} to efficiently reduce the overall search space of the split finding process by approximating the raw data distribution as a surrogate histogram representation. Empirically, it has been shown that quantile approximations work just as well as the exact greedy solutions \cite{chen2015xgboost,ke2017lightgbm,keck2017fastbdt}. This method for data approximations can serve as a semi-secure approach for training Gradient Boosted Decision Trees under a \textit{Trusted Federations} security policy. By quantizing or reducing the resolution of the raw distribution, we can effectively generate a surrogate representation of the raw data containing relatively lower fidelity of information than the original data distribution. Therefore, this does not directly reveal the raw data distribution of the original data source. 

Many methods for building distributed quantile sketches exist, including GKMethod \cite{greenwald2001space} and its extended variants \cite{zhang2007fast}. However, each comes with its trade-offs in performance, speed, and reconstruction accuracy. For our proposed method, this algorithm implements the Distributed Distribution Sketch or \textit{DDSketch} \cite{masson2019ddsketch}, an efficient and robust process that constructs highly accurate quantile sketch approximations of data distributions with the ability to merge multiple quantile sketches. Furthermore, another advantage of using \textit{DDSketch} enables training of XGBoost where parties can join during intermediate steps of the boosting process when a new party joins the federation. Due to its ability to merge quantile sketch histograms efficiently and accurately, this method enables dynamic adaptations to new data distributions in the data as new parties join the federation.

To train a gradient-boosted decision tree within an FL-setting, the aggregator first initializes a global null model, $f_{\emptyset}^{(\mathcal{A})}$ (line 1-2). The null model in this scenario is based on computing the average values of the target label from each of the local party's data distributions. For each party $p$, we obtain the sum of the target values as $Y_p$ and the corresponding sample size count, $n_p$ (Line 1). These values are collected from each local party and transmitted to the aggregator $\mathcal{A}$, where the aggregator computes the federation average of the target label values (Line 2).

Next, we then compute a federation-wise surrogate data representation that will be used in the boosting process when updating the tree model. To compute this surrogate data representation, we utilize a histogram-based method to compute approximations of the data. For each party $p$, we construct a data sketch of the local party's input features for a given dataset $\mathcal{D}_X^{(p_i)}$ (Line 5). For our implementation, we utilize DDSketch \cite{masson2019ddsketch} to construct the histogram representation for each of the party's local data distribution. During this process, we also provide a parameter $\delta$, which represents the relative error parameter used by the DDSketch algorithm. Controlling this parameter $\delta$ can dictate the amount of fidelity of the histogram used to approximate the data. Hence, the lower the error parameter used, the higher the risk of exposing the local party's data distribution. After computing the local party's histogram sketch representation, each party then transmits its histogram representation to the aggregator $\mathcal{A}$ (Line 6). When each of the parties finishes transmitting their local data distribution histograms to the aggregator, the aggregator then merges the histogram into a single histogram representation $\tilde{\mathcal{D}}^{\mathcal{A}}_X$.

After computing the surrogate histogram representation of the data, we then initiate the iterative federated learning process. First, the aggregator $\mathcal{A}$ transmits their global model, $f^{(\mathcal{A})}_t$ to each party, $p_i$, which is assigned to each party's respective local model $f^{(p_i)}_t$ (Line 13). We then evaluate $f_t^{(p_i)}$ on $\tilde{\mathcal{D}}^{(p_i)}_X$ to obtain the model's predictions, $\hat{y}_t^{(p_i)}$ (Line 14). Afterward, given the predictions, we compute the loss function which is used to compute the gradient, $g^{(p_i)}$, and Hessian's, $h^{(p_i)}$ for each of the corresponding surrogate input feature value split candidates (Line 15). Gradient and Hessian statistics that fall under a certain bin interval are grouped within their respective value buckets \cite{chen2015xgboost}. The gradient and Hessian for each party are then sent back as replies to the aggregator and then collected until some quorum criterion has been met (Line 16). Given the collected results from each party, we perform a fusion operation to merge the final histogram representation used towards boosting the decision tree model, as formulated in the method of DDSketch \cite{masson2019ddsketch} (Line 20).

Given the aggregated gradients $G^{(\mathcal{A})}_m$, Hessians, $H^{(\mathcal{A})}_m$, and the aggregated surrogate data representation histogram, $\tilde{\mathcal{D}}^{\mathcal{A}}_X$, we perform the boosting process to grow a new tree to derive a new global model, $f_t^{(\mathcal{A})}$ (Line 21). With a new $f_t^{(\mathcal{A})}$ generated, we repeat our training process for $T$ rounds, or until some other stopping criteria depending on whether early-stopping or other heuristics are considered (line 22).

\section{Experimental Setup}
In this section, we describe our experimental methodology for evaluating our proposed implementation of the Federated Gradient Boosting algorithm under various non-IID sample-wise data. We first introduce the various datasets used in our experiments. Then we describe the experimental setup for how we split our dataset into various partitions to evaluate our algorithm under different non-IID scenarios. Finally, we also present our method for how we evaluate our Federated Gradient Boosted Decision Tree-based model.

\subsection{Datasets}
\begin{table*}[]
\centering
\resizebox{\textwidth}{!}{%
\begin{tabular}{ccclccccclc}
\hline
\textbf{Dataset} & \textbf{Task Type} & \textbf{Partition   Variation} &  & \textbf{Party 1} & \textbf{Party 2} & \textbf{Party 3} & \textbf{Party 4} & \textbf{Party 5} &  & \textbf{Training Sample Size} \\ \hline
\multicolumn{1}{l}{} & \multicolumn{1}{l}{} & \multicolumn{1}{l}{} &  & \multicolumn{1}{l}{} & \multicolumn{1}{l}{} & \multicolumn{1}{l}{} & \multicolumn{1}{l}{} & \multicolumn{1}{l}{} &  & \multicolumn{1}{l}{} \\
\multirow{5}{*}{Bank} & \multirow{5}{*}{Binary} & Even Split &  & 7234 & 7234 & 7234 & 7233 & 7233 &  & \multirow{5}{*}{36168} \\
 &  & Partition A &  & 10851 & 9043 & 6329 & 6781 & 3164 &  &  \\
 &  & Partition B &  & 15373 & 9268 & 5820 & 4323 & 1384 &  &  \\
 &  & Partition C &  & 20007 & 8999 & 4157 & 2399 & 606 &  &  \\
 &  & Partition D &  & 24507 & 7617 & 2538 & 1241 & 265 &  &  \\
\multicolumn{1}{l}{} & \multicolumn{1}{l}{} & \multicolumn{1}{l}{} &  & \multicolumn{1}{l}{} & \multicolumn{1}{l}{} & \multicolumn{1}{l}{} & \multicolumn{1}{l}{} & \multicolumn{1}{l}{} &  & \multicolumn{1}{l}{} \\
\multirow{5}{*}{Bitcoin} & \multirow{5}{*}{Multiclass} & Even Split &  & 420927 & 420927 & 420926 & 420926 & 420926 &  & \multirow{5}{*}{2104632} \\
 &  & Partition A &  & 631391 & 526157 & 368311 & 394618 & 184155 &  &  \\
 &  & Partition B &  & 894469 & 539312 & 338714 & 251569 & 80568 &  &  \\
 &  & Partition C &  & 1164125 & 523692 & 241909 & 139658 & 35248 &  &  \\
 &  & Partition D &  & 1425971 & 443278 & 147763 & 72199 & 15421 &  &  \\
\multicolumn{1}{l}{} & \multicolumn{1}{l}{} & \multicolumn{1}{l}{} &  & \multicolumn{1}{l}{} & \multicolumn{1}{l}{} & \multicolumn{1}{l}{} & \multicolumn{1}{l}{} & \multicolumn{1}{l}{} &  & \multicolumn{1}{l}{} \\
\multirow{5}{*}{Credit Card} & \multirow{5}{*}{Binary} & Even Split &  & 45569 & 45569 & 45569 & 45569 & 45569 &  & \multirow{5}{*}{227845} \\
 &  & Partition A &  & 68353 & 56962 & 39873 & 42721 & 19936 &  &  \\
 &  & Partition B &  & 96834 & 58386 & 36669 & 27234 & 8722 &  &  \\
 &  & Partition C &  & 126027 & 56695 & 26188 & 15119 & 3816 &  &  \\
 &  & Partition D &  & 154375 & 47988 & 15996 & 7816 & 1670 &  &  \\
\multicolumn{1}{l}{} & \multicolumn{1}{l}{} & \multicolumn{1}{l}{} &  & \multicolumn{1}{l}{} & \multicolumn{1}{l}{} & \multicolumn{1}{l}{} & \multicolumn{1}{l}{} & \multicolumn{1}{l}{} &  & \multicolumn{1}{l}{} \\
\multirow{5}{*}{Dry Bean} & \multirow{5}{*}{Multiclass} & Even Split &  & 2178 & 2178 & 2178 & 2177 & 2177 &  & \multirow{5}{*}{10888} \\
 &  & Partition A &  & 3267 & 2723 & 1905 & 2041 & 952 &  &  \\
 &  & Partition B &  & 4629 & 2790 & 1752 & 1301 & 416 &  &  \\
 &  & Partition C &  & 6024 & 2709 & 1251 & 722 & 182 &  &  \\
 &  & Partition D &  & 7378 & 2293 & 764 & 373 & 80 &  &  \\
\multicolumn{1}{l}{} & \multicolumn{1}{l}{} & \multicolumn{1}{l}{} &  & \multicolumn{1}{l}{} & \multicolumn{1}{l}{} & \multicolumn{1}{l}{} & \multicolumn{1}{l}{} & \multicolumn{1}{l}{} &  & \multicolumn{1}{l}{} \\
\multirow{5}{*}{HTRU 2} & \multirow{5}{*}{Binary} & Even Split &  & 2864 & 2864 & 2864 & 2863 & 2863 &  & \multirow{5}{*}{14318} \\
 &  & Partition A &  & 4296 & 3580 & 2505 & 2684 & 1253 &  &  \\
 &  & Partition B &  & 6086 & 3669 & 2304 & 1711 & 548 &  &  \\
 &  & Partition C &  & 7920 & 3563 & 1645 & 950 & 240 &  &  \\
 &  & Partition D &  & 9702 & 3015 & 1005 & 491 & 105 &  &  \\
\multicolumn{1}{l}{} & \multicolumn{1}{l}{} & \multicolumn{1}{l}{} &  & \multicolumn{1}{l}{} & \multicolumn{1}{l}{} & \multicolumn{1}{l}{} & \multicolumn{1}{l}{} & \multicolumn{1}{l}{} &  & \multicolumn{1}{l}{} \\
\multirow{5}{*}{Firewall} & \multirow{5}{*}{Multiclass} & Even Split &  & 10485 & 10485 & 10485 & 10485 & 10485 &  & \multirow{5}{*}{52425} \\
 &  & Partition A &  & 15727 & 13107 & 9174 & 9830 & 4587 &  &  \\
 &  & Partition B &  & 22281 & 13433 & 8438 & 6266 & 2007 &  &  \\
 &  & Partition C &  & 28997 & 13045 & 6026 & 3479 & 878 &  &  \\
 &  & Partition D &  & -- & -- & -- & -- & -- &  &  \\
\multicolumn{1}{l}{} & \multicolumn{1}{l}{} & \multicolumn{1}{l}{} &  & \multicolumn{1}{l}{} & \multicolumn{1}{l}{} & \multicolumn{1}{l}{} & \multicolumn{1}{l}{} & \multicolumn{1}{l}{} &  & \multicolumn{1}{l}{} \\
\multirow{5}{*}{Parkinson} & \multirow{5}{*}{Binary} & Even Split &  & 121 & 121 & 121 & 121 & 120 &  & \multirow{5}{*}{604} \\
 &  & Partition A &  & 181 & 152 & 106 & 113 & 52 &  &  \\
 &  & Partition B &  & 257 & 156 & 97 & 71 & 23 &  &  \\
 &  & Partition C &  & 335 & 151 & 68 & 40 & 10 &  &  \\
 &  & Partition D &  & 411 & 126 & 42 & 21 & 4 &  &  \\
\multicolumn{1}{l}{} & \multicolumn{1}{l}{} & \multicolumn{1}{l}{} &  & \multicolumn{1}{l}{} & \multicolumn{1}{l}{} & \multicolumn{1}{l}{} & \multicolumn{1}{l}{} & \multicolumn{1}{l}{} &  & \multicolumn{1}{l}{} \\ \hline
\multicolumn{1}{l}{} & \multicolumn{1}{l}{} & \multicolumn{1}{l}{} &  & \multicolumn{1}{l}{} & \multicolumn{1}{l}{} & \multicolumn{1}{l}{} & \multicolumn{1}{l}{} & \multicolumn{1}{l}{} &  & \multicolumn{1}{l}{} \\
\end{tabular}%
}
\caption{Sample Counts of Each Party for Every Dataset and Corresponding Partition Scenarios}
\label{tab:my-table2}
\end{table*}

We evaluate our algorithm with 7 different datasets ranging from Kaggle \cite{kaggle} and the University of California, Irvine Machine Learning Repository (UCI ML Repository) \cite{Dua:2019}. All datasets are classification task-based and are tabular data. They are a mix of binary and multiclass classification tasks and have a varying number of instances. We consider datasets of skewed labels and differing numbers of features, spanning from 9 to 756. The datasets span multiple subjects including science, finance, healthcare, and miscellaneous areas, which allows us to evaluate various non-IID scenarios. The entire list of the dataset utilized in this experiment is presented in Table \ref{tab:my-table}.

The Bank Marketing Dataset (\textbf{Bank}) was taken from the UCI ML Repository \cite{MORO201422}. The focus of this dataset is classifying whether a customer subscribed to a marketing campaign done by a bank based on the characteristics of the customer and the specifics of the campaign. These 20 characteristics make up the feature values and included how many times and day of the week the customer was contacted. There are 45,211 instances in this dataset and is formulated as a binary classification task.

Bitcoin Heist Ransomware Address Dataset (\textbf{Bitcoin}) is sourced also from the UCI ML Repository \cite{akcora2019bitcoinheist}. This dataset has features on executed bitcoin transactions, such as information on the merging pattern and if the coins were split. It classifies if the transaction was hit with ransomware, and if so, classifies the type of ransomware. This dataset is significantly larger, originally with 2,916,697 instances, but was preprocessed to only include one transaction per originating address. There were also originally many types of ransomware, but only the top 6 are utilized. This was performed since the aggregator needs all the parties to have the same number of classes and many ransomware only have single instances.

Credit Card Fraud Dataset (\textbf{Credit Card}) is taken from Kaggle \cite{KaggleCredit}. The dataset was preprocessed through PCA from the original data from transactions to preserve the privacy of the original dataset. The dataset contains 284,807 instances with only 492 instances being classified as a fraudulent transactions, making this a binary classification task. This makes the dataset highly imbalanced, with less than 1\% of the whole dataset being in a positive class.

We use an already encoded image classification dataset, which is the \textbf{Dry Bean} Dataset from the UCI ML Repository \cite{KOKLU2020105507}. The encoding done makes up the features of the dataset as characteristics of the beans, such as area. It has 13,611 instances and is based on a multiclass classification task with 7 labels.

The \textbf{HTRU 2} Dataset is also sourced from the UCI ML Repository and is a binary dataset with 9 features \cite{10.1093/mnras/stw656}. It determines whether a pulsar candidate is a pulsar and is imbalanced with 1,639 of the 17,898 instances classified as positive. The features all pertain to statistics of the integrated pulse profile or the DM-SNR curve of the pulsar.

The Internet Firewall Dataset (\textbf{Firewall}) is a multiclass dataset of whether an action was allowed or not by a firewall system \cite{8355382}. It is taken from the UCI ML Repository and has 65,532 instances spread across 4 labels. With features pertaining to the transaction like source and destination port. This dataset only has experiments run on 3 of the partitions since the fourth would create a party with less than 4 labels, and thus could not be run on the Federated Learning framework. To run, all parties need to have data with the same number of labels.

The final dataset is the \textbf{Parkinson’s} Dataset which is sourced from UCI ML Repository and is binary of whether a patient has Parkinson’s \cite{SAKAR2019255}. This is based on 754 features, which makes an interesting problem for the algorithm since as the number of features increases, so does the complexity of the decision trees thus increasing the time spent on each iteration. This dataset is 756 instances in size.

The datasets have limited preprocessing done other than encoding the data into numeric values or deleting non-unique values.

\subsection{Party Setup and Configurations}
Models are evaluated using a 5-party Federated Learning setup. Each party has access to one partition of data as well as an evaluation dataset which is then shared with the aggregator to evaluate the model created by the partitions of data. Each partition contributes to the training dataset used by the aggregator and has access to the test dataset used.

We run experiments with $\epsilon=255$ max bins for the response histograms and with 100 max rounds. For the DDSketch relative error parameter we utilize a value of $\delta = 0.01$. We tune hyperparameters using an l2-regularization of 0.1 and a learning rate also of 0.1. The experiments are run using the IBM Federated Learning framework \cite{https://doi.org/10.48550/arxiv.2007.10987}.

\subsection{Dataset Split}
\begin{table}[]
\centering
\resizebox{\columnwidth}{!}{%
\begin{tabular}{cccccc}
\hline
\textbf{Partition} & \textbf{Party 1} & \textbf{Party 2} & \textbf{Party 3} & \textbf{Party 4} & \textbf{Party 5} \\ \hline
Even Split & 0.20 & 0.20 & 0.20 & 0.20 & 0.20 \\
Partition A & 0.30 & 0.25 & 0.17 & 0.19 & 0.09 \\
Partition B & 0.43 & 0.26 & 0.16 & 0.12 & 0.04 \\
Partition C & 0.55 & 0.25 & 0.11 & 0.07 & 0.02 \\
Partition D & 0.68 & 0.21 & 0.07 & 0.03 & 0.01 \\ \hline
\multicolumn{1}{l}{} & \multicolumn{1}{l}{} & \multicolumn{1}{l}{} & \multicolumn{1}{l}{} & \multicolumn{1}{l}{} & \multicolumn{1}{l}{} \\
\end{tabular}%
}
\caption{Ratio values of each party for different non-IID data partition}
\label{tab:my-table}
\end{table}

We split datasets to simulate non-IID cases by reallocating some samples that each party has, thus becoming increasingly imbalanced. This heterogeneity in data was done to evaluate how the model would adapt to distribution changes in the parties. Real-world applications of the framework will use parties that have data distributed differently across them, in most cases. Thus the framework will need to have consistent results despite the heterogeneity of the parties’ data.

These partitions are all done using 80\% of the data and the rest is preserved for evaluation. First, starting with each party having an equal number of samples as Sample Even experiments. Then move on to creating Sample Split A, which is done by reallocating 50\% the number of instances from Party 2 and moving it to Party 1. Then 75\% of the instances from Party 3 and moving it to Party 2. This was similarly done for Party 4 and 5, where 62.5\% was moved to Party 3 and where 56.25\% was moved to Party 4 respectively. This creates an unequal distribution of the number of instances for each party. This process was repeated 3 more times for all datasets, except for the firewall dataset which was only repeated twice, creating an increasing imbalanced distribution of instances for the parties. Thus, in the end, Party 1 has 68\% of the number of instances in the dataset, and Party 5 has less than 1\%. The ratio of each of the parties, relative to the whole training set, for each of the partitions, is shown in Table \ref{tab:my-table}. The actual size of each party used for each reallocation for each dataset is noted in Table \ref{tab:my-table2}.

This method of reallocating the data should keep the label distribution of each of the parties consistent with the original dataset as well as each other party thus label non-IID is not an underlying factor for influencing the results. We perform such data reallocation, as the data is randomly sampled, the sample taken should continue to follow the same distribution as the original population. Thus the instances that are reallocated to the previous party should hold the same label and feature distribution as the originating party and the party it is being added to. Additionally, once the reallocated instances are added to the party, the label and feature distribution should not change for the originating or new party based on the loss or addition of instances.

\subsection{Evaluation Methodology}
The partitions of data are evaluated against both a centralized case as well as against the local parties. The centralized case is done by using a global holdout set of 20\% of the data to evaluate the performance of the model that uses the remainder of the data, or 80\%. The local parties are each party’s data for every partition, so the five from Sample Even and five from Sample Split A, and so on, are evaluated against the same holdout set as before. The local party results are averaged for each of the datasets. 

We evaluate the partitions using 5 parties and the same global holdout set as previous evaluations to test model performance with the non-IID data. Thus, the aggregator receives responses from five parties to build the model, instead of one which the previous evaluations use. We use the F1 score for the evaluation metric of all the models and compare the centralized model as well as each of the partitions.

\section{Results}
\begin{table*}[ht!]
\tiny
\centering
\resizebox{\linewidth}{!}{%
\begin{tabular}{lccccccc}
\hline
 & \textbf{Bank} & \textbf{Bitcoin} & \textbf{Credit Card} & \textbf{Dry Bean} & \textbf{HTRU 2} & \textbf{Firewall} & \textbf{Parkinson} \\ \hline
Centralized & 0.56 & 0.29 & 0.82 & 0.93 & 0.89 & 0.77 & 0.93 \\
Local Party & 0.49 & 0.25 & 0.79 & 0.90 & 0.87 & 0.76 & 0.86 \\ \hline
Sample Even & 0.56 & 0.27 & 0.80 & 0.93 & 0.89 & 0.77 & 0.93 \\
Sample Split A & 0.56 & 0.27 & 0.78 & 0.93 & 0.89 & 0.77 & 0.93 \\
Sample Split B & 0.56 & 0.29 & 0.79 & 0.93 & 0.89 & 0.77 & 0.93 \\
Sample Split C & 0.56 & 0.28 & 0.83 & 0.93 & 0.89 & 0.77 & 0.93 \\
Sample Split D & 0.56 & 0.27 & 0.84 & 0.93 & 0.89 & -- & 0.93 \\ \hline
\multicolumn{1}{l}{} & \multicolumn{1}{l}{} & \multicolumn{1}{l}{} & \multicolumn{1}{l}{} & \multicolumn{1}{l}{} & \multicolumn{1}{l}{} & \multicolumn{1}{l}{} & \multicolumn{1}{l}{} \\
\end{tabular}%
}
\caption{F1 Score of Experiments for Federated XGBoost Under Various Non-IID Partition Scenarios}
\label{tab:my-table3}
\end{table*}

In this section, we explain the results found from running experiments. The results for all datasets and different partitions used are shown in Table \ref{tab:my-table3}. It shows the partitions’ F1 score in comparison to the centralized cases and the local party. Despite the reallocation of data to different parties, the experiments continued to meet the performance of the centralized model and most exceed that of the local parties. For most of the datasets, less the Bitcoin and Credit Card datasets, and despite the changes in allocation, the F1 score is consistent across all reallocations to the hundredths decimal point. Despite even the most extreme reallocations, the framework maintains consistency with the results in the changing partitions.

For the Bitcoin and Credit Card datasets, the F1 scores are not consistent across the different partitions. They are the 2 largest datasets used with 2,104,632 and 227,845 instances in their training datasets, respectively. Thus the discrepancies could be due to each of the parties still being relatively large after reallocation, especially compared to the other datasets. Differences in the results from these 2 datasets are the specific partitions with different F1 scores comparatively to the rest of the partitions as well as some of the F1 scores for the Credit Card dataset being less than or equal to the local parties. As shown by the other datasets' experiments, the partition results tend to be consistent with the centralized case. Thus, the Bitcoin dataset's results mostly being 0.27 is irregular since it does not match the centralized case, but shows consistency across 3 of the parties. Sample Split B and C being different could be due to the label distribution, which is generally consistent with the original dataset but could have some slight variation, or simply the size of the dataset causing variability in the results from the framework. Focusing on the Credit Card dataset, none of the results are consistent with either the centralized case or the other partitions. It is interesting to note that while the Bitcoin dataset showed variation in Sample Split B and C, the Credit Card dataset seems to show the most variation in Sample Split A and B.  But what is different than the Bitcoin dataset is that the variable partitions have a lower F1 score than the rest of the reallocations. They are lower than the local parties which should not occur. The local parties are averages of all parties from all partitions being trained and tested and thus do not benefit from the aggregation of multiple models created during the training. Thus, the local parties should be lower than the sample split's results. This could be due to the label distribution or the size, or specifically for this dataset the highly unbalanced nature of the dataset.
\section{Conclusion}

In this paper, we propose a Federated XGBoost implementation for use on sample-wise non-IID data. In the approach, we use surrogate histograms of the data distributions from the different parties to create an aggregated model. This is done to preserve the privacy of the parties' data while still creating an accurate model using the XGBoost algorithm. We focus on testing the implementation against sample-wise non-IID data to see how the algorithm reacts to the change in data distribution. In our experiments, we see despite the changing size of the party's datasets (done by creating sample-wise non-IID data) we continue to see consistent results from the different partitioning of data. In most cases, the sample splits' F1 score meets the centralized case and exceeds the local parties' results. The consistency makes this Federated Learning XGBoost framework suitable to handle Federated Learning cases where XGBoost is the most appropriate algorithm. In the future, we will test the same Federated XGBoost on Label non-IID and Feature non-IID. That would allow our algorithm to be used for any sort of non-IID or heterogeneous data moving forward.

In future explorations on evaluating the limitations of our model against other non-IID scenarios, we will consider other variants of non-IID data distribution scenarios. This includes non-IID data against feature-wise skews, target-label skews, and other types of non-IID distribution characteristics that may impact the model's performance. These studies would further our understanding of how different types of non-IID data can potentially degrade the performance of Federated Gradient Boosted Decision Tree-based algorithms. 

\bibliographystyle{IEEEtran}
\bibliographystyle{chicago}
\bibliography{References.bib}
\vspace{12pt}

\end{document}